\documentclass[10pt,twocolumn,letterpaper]{article}

\usepackage{cvpr}      % To produce the REVIEW version
\usepackage{graphicx}
\usepackage{amsmath}
\usepackage{amssymb}
\usepackage{booktabs}
\usepackage[document]{ragged2e}
\usepackage[T1]{fontenc}
\usepackage[pagebackref,breaklinks,colorlinks]{hyperref}

\usepackage[capitalize]{cleveref}
\crefname{section}{Sec.}{Secs.}
\Crefname{section}{Section}{Sections}
\Crefname{table}{Table}{Tables}
\crefname{table}{Tab.}{Tabs.}

\def\cvprPaperID{*****} % *** Enter the CVPR Paper ID here
\def\confName{CVPR}
\def\confYear{2023}

\begin{document}

\title{Exploring the Lottery Ticket Hypothesis with Explainability Methods: Insights into Sparse Network Performance}

\author{Shantanu Ghosh,  Kayhan Batmanghelich\\
Department of Electrical and Computer Engineering, Boston University\\
8 St Mary's St, Boston, MA 02215
\\
{\tt\small \{shawn24, batman\}@bu.edu}
}
\maketitle

%%%%%%%%% ABSTRACT
\begin{abstract}
   Discovering a high-performing sparse network within a massive neural network is advantageous for deploying them on devices with limited storage, such as mobile phones. 
Additionally, model explainability is essential to fostering trust in AI.
The Lottery Ticket Hypothesis (LTH) finds a network within a deep network with comparable or superior performance to the original model. However, limited study has been conducted on the success or failure of LTH in terms of explainability.
In this work, we examine why the performance of the pruned networks gradually increases or decreases. 
Using Grad-CAM and Post-hoc concept bottleneck models (PCBMs), respectively, we investigate the explainability of pruned networks in terms of pixels and high-level concepts.
We perform extensive experiments across vision and medical imaging datasets. As more weights are pruned, the performance of the network degrades. The discovered concepts and pixels from the pruned networks are inconsistent with the original network -- a possible reason for the drop in performance.
\end{abstract}

%%%%%%%%% BODY TEXT
%%%%%%%%% BODY TEXT
\section{Introduction}
Neural network pruning~\cite{lecun1989optimal, han2015learning, li2016pruning} removes irrelevant parameters to optimize storage requirements, reduce energy consumption, and perform efficient inference. The Lottery Ticket Hypothesis (LTH)~\cite{frankle2018lottery} finds a subnetwork within a deep network by pruning the superfluous weights based on their magnitudes. Model explainability is important to engender trust in prediction. However, little exhaustive research on the explainability of LTH has been conducted. In this paper, we  investigate  the improvement/decline in pruned networks using LTH by analyzing whether they rely on relevant pixels / interpretable concepts for prediction. We accomplish this by quantifying local (explaining an individual sample) and global explanations (explaining a class) from the pruned networks using Grad-CAM-based saliency maps and PCBMs, respectively.

The literature of explaining a network in terms of pixels is quite extensive. The methods such as model attribution (\eg Saliency Map~\cite{simonyan2013deep, selvaraju2016grad}), counterfactual approach ~\cite{abid2021meaningfully, singla2019explanation}, and distillation methods~\cite{alharbi2021learning, cheng2020explaining} are examples of post hoc explainability approaches. Those methods either identify important features of input that contribute the most to the network's output~\cite{shrikumar2016not}, generate perturbation to the input that flips the network's output~\cite{samek2016evaluating}, \cite{montavon2018methods}, or estimate simpler functions that locally approximate the network output. 
Later~\cite{koh2020concept} proposes an interpretable-by-design concept bottleneck model (CBM) , in which they first identify the human interpretable concepts from the images and then utilize the concepts to predict the labels using an interpretable classifier.~\cite{yuksekgonul2022post} learns the concepts from the embedding of a trained model in PCBM. Recently, ~\cite{ghosh2023route} carves a set of interpretable models from a trained model.
However, little emphasis has been given to evaluating the explanations from the networks obtained by network pruning. In~\cite{frankle2019dissecting}, the authors discover the neuron-concept relationship by applying Net-dissection~\cite{bau2017network}. They did not study whether the set of discriminating concepts in pruned networks remains the same or changes when networks are pruned.

\begin{figure*}[t]
\centering
\includegraphics[width=1\linewidth]{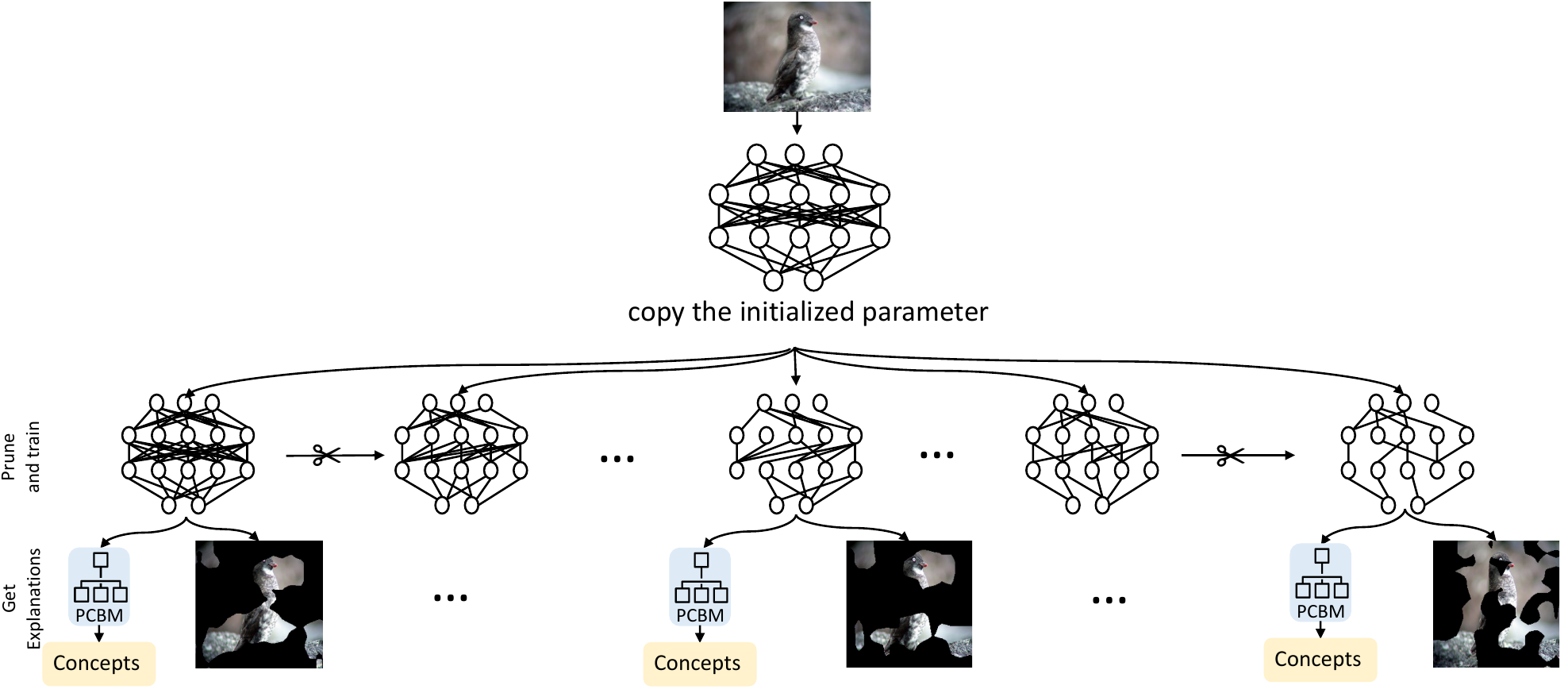}
\caption{Overview of our method. (1) First, we prune a deep neural network using \emph{Lottery Ticket Hypothesis}~\cite{frankle2018lottery}. (2) For each pruned subnetwork, we identify the top human interpretable concepts for classification using PCBM~\cite{yuksekgonul2022post}. (3) For each pruned subnetwork, we compute the local explanation using Grad-CAM~\cite{selvaraju2016grad}.}
\label{fig:schematic}
\end{figure*}

In this paper, we study the relationship between pruning and explainability. Initially, we prune the deep model using LTH. We then test either of the following hypotheses:
\noindent\textbf{Hypothesis A.} The pruned networks prioritize the same relevant concepts/pixels for prediction as the original network. Hence pruning has no effect on the global/local explanations. As a result, LTH is able to identify the \emph{winning tickets}, i.e. subnetworks with comparable or superior performance to the original network.
\noindent\textbf{Hypothesis B.}  Pruning modifies the global/local explanations as the pruned networks prioritize different concepts/pixels for prediction compared to the original network. Consequently, LTH is unable to find the \emph{winning tickets}. To validate the two hypotheses, we borrow tools from explainable AI to quantify explanations for different pruned networks. Specifically, we use Grad-CAM to estimate the local explanations and use PCBM to identify the important concepts from the embeddings of each pruned network. To our knowledge, we are the first to investigate the concept-based explainability of LTH using PCBM.
% TODO by Shantanu:
% Mention 3 hypothesis by Bau et al.
% define Global vs local explanation

\section{Explaining Lottery Ticket Hypothesis}
\noindent\textbf{Overview.}
We train a neural network $\displaystyle f:\mathcal{X} \rightarrow \mathcal{Y}$, taking images $\mathcal{X}$ as input to predict the labels $\mathcal{Y} \in \mathbb{R}^K$, $K$ being the number of classes. 
% We denote the output logit for the $k^{th}$ class label and image $x$ as $f_k(x) = h_k(\Phi(x)$. 
We assume that $\displaystyle f$ is a composition $\displaystyle h \circ \Phi $, where $\displaystyle \Phi: \mathcal{X} \rightarrow \mathbb{R}^l $ is the  image embedding and $\displaystyle h: \mathbb{R}^l \rightarrow \mathcal{Y}$ is a classifier. 
Figure \ref{fig:schematic} shows an overview of our approach.
% We denote the binary classifier $\displaystyle t_C: \mathbb{R}^l \rightarrow\mathbb{R}$ aiming to estimate a unit linear CAV $v_C \in \mathbb{R}^l$ as for the concept $C$, orthogonal to the classification boundary.
\subsection{Pruning Methodology} 
\emph{LTH}~\cite{frankle2018lottery} aims to find a pruned subnetwork within a deep neural network that achieves similar accuracy as the original network -- \emph{when trained in isolation}. These pruned networks are finetuned for a small number of iterations to achieve accuracy. 
 We prune and fine-tune the network $f$ simultaneously for $n$ rounds using the strategy of iterative magnitude pruning in \emph{LTH}~\cite{frankle2018lottery}. Thus we obtain $f^1, f^2, \dots, f^n$ subnetworks where the network $f$ for the $i^{th}$ round is denoted as $f^i(.) = h^i(\Phi^i(.))$. 
 For brevity, we drop \emph{round index} $i$ from the notations in the following sections.

\begin{figure}
\centering
\includegraphics[width=1\linewidth]{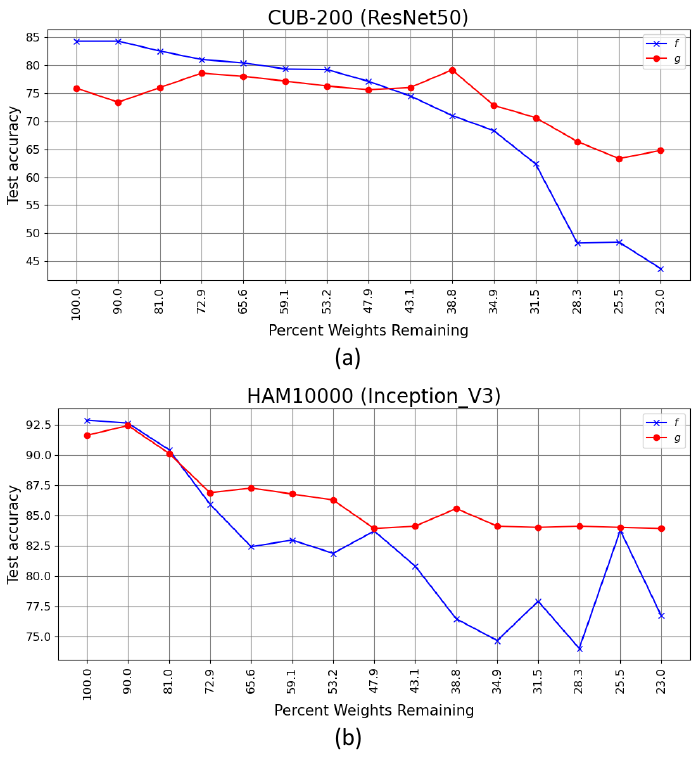}
\caption{The test accuracy on CUB-200 and HAM10000 when the network is pruned to the specified size.}
\label{fig:quant_accuracy}
\end{figure}

\begin{table*}[t]
\caption{The top-3 concepts with the highest weights of the carved interpretable models ($g$) using PCBM from the pruned subnetworks for various pruning iterations for the bird species of CUB-200 dataset. Due to the space constraint, we only report the concepts from $g$ extracted from the model for iterations 1, 2, 4, and 15 with 100\%, 90\% 72.9\%, and 23\% weights remaining.}
\fontsize{6.2pt}{0.20cm}\selectfont
\label{tab:cub}
\begin{center}
\begin{tabular}{l l l l  l}
\toprule 
        \textbf{LABEL} 
       & \textbf{100.0 \% WEIGHTS REMAINING} 
       & \textbf{90.0 \% WEIGHTS REMAINING} 
       % & \textbf{81.0 \% WEIGHTS REMAINING} 
       & \textbf{72.9 \% WEIGHTS REMAINING} 
       % & \textbf{25.5 \% WEIGHTS REMAINING} 
       & \textbf{23.0 \% WEIGHTS REMAINING} \\
\midrule 
    Black footed Albatross & 
    \vtop{
        \hbox{\strut 1. \textbf{bill\_shape\_hooked\_seabird}}
        \hbox{\strut 2. \textbf{under\_tail\_color\_black}}
        \hbox{\strut 3. \textbf{size\_medium\_9\_16\_in}}
    }& 
    \vtop{
        \hbox{\strut 1. \textbf{bill\_shape\_hooked\_seabird}}
        \hbox{\strut 2. \textbf{under\_tail\_color\_black}}
        \hbox{\strut 3. wing\_pattern\_solid}
    }
    % & 
    % \vtop{
    %     \hbox{\strut 1. \textbf{bill\_shape\_hooked\_seabird}}
    %     \hbox{\strut 2. wing\_pattern\_solid}
    %     \hbox{\strut 3. \textbf{size\_medium\_9\_16\_in}}
    % }
    & 
    \vtop{
        \hbox{\strut 1. \textbf{bill\_shape\_hooked\_seabird}}
        \hbox{\strut 2. leg\_color\_grey}
        \hbox{\strut 3. \textbf{size\_medium\_9\_16\_in}}
    }
    % & 
    % \vtop{
    %     \hbox{\strut 1. \textbf{bill\_shape\_hooked\_seabird}}
    %     \hbox{\strut 2. crown\_color\_white}
    %     \hbox{\strut 3.  \textbf{size\_medium\_9\_16\_in}}
    % }
    & 
    \vtop{
        \hbox{\strut 1. \textbf{bill\_shape\_hooked\_seabird}}
        \hbox{\strut 2. underparts\_color\_grey}
        \hbox{\strut 3. wing\_pattern\_solid}
    }\\  
    
\midrule    
Rusty Blackbird &\vtop{
        \hbox{\strut 1. \textbf{back\_color\_brown}}
        \hbox{\strut 2. \textbf{breast\_color\_brown}}
        \hbox{\strut 3. \textbf{wing\_color\_black}}
    }& 
    \vtop{
        \hbox{\strut 1. underparts\_color\_buff}
        \hbox{\strut 2. \textbf{breast\_color\_brown}}
        \hbox{\strut 3. \textbf{back\_color\_brown}}
    }
    % & 
    % \vtop{
    %     \hbox{\strut 1. \textbf{breast\_color\_brown}}
    %     \hbox{\strut 2. \textbf{back\_color\_brown}}
    %     \hbox{\strut 3. bill\_shape\_allpurpose}
    % }
    & 
    \vtop{
        \hbox{\strut 1. \textbf{breast\_color\_brown}}
        \hbox{\strut 2. belly\_color\_grey}
        \hbox{\strut 3. underparts\_color\_grey}
    }
    % & 
    % \vtop{
    %     \hbox{\strut 1. \textbf{back\_color\_brown}}
    %     \hbox{\strut 2. bill\_shape\_hooked\_seabird}
    %     \hbox{\strut 3. bill\_shape\_allpurpose}
    % }
    & 
    \vtop{
        \hbox{\strut 1. underparts\_color\_buff}
        \hbox{\strut 2. \textbf{back\_color\_brown}}
        \hbox{\strut 3. belly\_color\_grey}
    } \\
\midrule    
White Pelican &\vtop{
        \hbox{\strut 1. \textbf{bill\_shape\_dagger}}
        \hbox{\strut 2. \textbf{upper\_tail\_color\_white}}
        \hbox{\strut 3. \textbf{back\_pattern\_multicolored}}
    }&  
    \vtop{
        \hbox{\strut 1. wing\_pattern\_solid}
        \hbox{\strut 2. bill\_color\_grey}
        \hbox{\strut 3. \textbf{upper\_tail\_color\_white}}
    }
    % & 
    % \vtop{
    %     \hbox{\strut 1. underparts\_color\_yellow}
    %     \hbox{\strut 2. \textbf{upper\_tail\_color\_white}}
    %     \hbox{\strut 3. leg\_color\_grey}
    % }
    &
    \vtop{
        \hbox{\strut 1. belly\_color\_yellow}
        \hbox{\strut 2. primary\_color\_yellow}
        \hbox{\strut 3. underparts\_color\_yellow}
    }
    % &
    % \vtop{
    %     \hbox{\strut 1. wing\_pattern\_multicolored}
    %     \hbox{\strut 2. under\_tail\_color\_grey}
    %     \hbox{\strut 3. belly\_pattern\_solid}
    % }
    & 
    \vtop{
        \hbox{\strut 1. wing\_pattern\_multicolored}
        \hbox{\strut 2. bill\_color\_grey}
        \hbox{\strut 3. wing\_color\_white}
    } \\ 
\midrule
Grasshopper Sparrow &\vtop{
        \hbox{\strut 1. \textbf{back\_color\_buff}}
        \hbox{\strut 2. \textbf{wing\_pattern\_striped}}
        \hbox{\strut 3. \textbf{back\_pattern\_striped}}
    }& 
    \vtop{
        \hbox{\strut 1. \textbf{back\_color\_buff}}
        \hbox{\strut 2. \textbf{wing\_pattern\_striped}}
        \hbox{\strut 3. belly\_color\_buff}
    }
    % & 
    % \vtop{
    %     \hbox{\strut 1. \textbf{back\_color\_buff}}
    %     \hbox{\strut 2. wing\_color\_white}
    %     \hbox{\strut 3. upperparts\_color\_buff}
    % }
    & 
    \vtop{
        \hbox{\strut 1. belly\_color\_buff}
        \hbox{\strut 2. wing\_color\_white}
        \hbox{\strut 3. upperparts\_color\_buff}
    }
    % & 
    % \vtop{
    %     \hbox{\strut 1. breast\_color\_buff}
    %     \hbox{\strut 2. primary\_color\_buff}
    %     \hbox{\strut 3. belly\_pattern\_solid}
    % }
    & 
    \vtop{
        \hbox{\strut 1. \textbf{back\_color\_buff}}
        \hbox{\strut 2. belly\_color\_buff}
        \hbox{\strut 3. underparts\_color\_buff}
    } \\
\midrule 
Vesper Sparrow &\vtop{
        \hbox{\strut 1. \textbf{leg\_color\_black}}
        \hbox{\strut 2. \textbf{nape\_color\_buff}}
        \hbox{\strut 3. \textbf{primary\_color\_buff}}
    }& 
    \vtop{
        \hbox{\strut 1. \textbf{nape\_color\_buff}}
        \hbox{\strut 2. \textbf{primary\_color\_buff}}
        \hbox{\strut 3. head\_pattern\_plain}
    }
    % &
    % \vtop{
    %     \hbox{\strut 1. \textbf{nape\_color\_buff}}
    %     \hbox{\strut 2. leg\_color\_black}
    %     \hbox{\strut 3. forehead\_color\_brown}
    % }
    & 
    \vtop{
        \hbox{\strut 1. \textbf{nape\_color\_buff}}
        \hbox{\strut 2. back\_color\_buff}
        \hbox{\strut 3. wing\_color\_buff'}
    }
    % &
    % \vtop{
    %     \hbox{\strut 1. breast\_pattern\_striped}
    %     \hbox{\strut 2. \textbf{nape\_color\_buff}}
    %     \hbox{\strut 3. breast\_color\_black}
    % }
    & 
    \vtop{
        \hbox{\strut 1. under\_tail\_color\_buff}
        \hbox{\strut 2. breast\_color\_black}
        \hbox{\strut 3. wing\_color\_buff}
    } \\
\midrule
Heermann Gull &\vtop{
        \hbox{\strut 1. \textbf{breast\_color\_grey}}
        \hbox{\strut 2. \textbf{upper\_tail\_color\_black}}
        \hbox{\strut 3. \textbf{underparts\_color\_grey}}
    }& 
    \vtop{
        \hbox{\strut 1. \textbf{breast\_color\_grey}}
        \hbox{\strut 2. \textbf{underparts\_color\_grey}}
        \hbox{\strut 3. throat\_color\_grey}
    }
    % &
    % \vtop{
    %     \hbox{\strut 1. \textbf{breast\_color\_grey}}
    %     \hbox{\strut 2. belly\_color\_grey}
    %     \hbox{\strut 3. leg\_color\_black}
    % }
    & 
    \vtop{
        \hbox{\strut 1. belly\_color\_grey}
        \hbox{\strut 2. \textbf{breast\_color\_grey}}
        \hbox{\strut 3. \textbf{underparts\_color\_grey}}
    }
    % &
    % \vtop{
    %     \hbox{\strut 1. head\_pattern\_plain}
    %     \hbox{\strut 2. primary\_color\_grey}
    %     \hbox{\strut 3. wing\_pattern\_solid}
    % }
    & 
    \vtop{
        \hbox{\strut 1. head\_pattern\_plain}
        \hbox{\strut 2. back\_color\_grey}
        \hbox{\strut 3. primary\_color\_grey}
    } \\ 
\midrule
American Crow &\vtop{
        \hbox{\strut 1. \textbf{crown\_color\_grey}}
        \hbox{\strut 2. \textbf{forehead\_color\_grey}}
        \hbox{\strut 3. \textbf{wing\_shape\_roundedwings}}
    }& 
    \vtop{
        \hbox{\strut 1. \textbf{wing\_shape\_roundedwings}}
        \hbox{\strut 2. \textbf{forehead\_color\_grey}}
        \hbox{\strut 3. primary\_color\_grey}
    }
    % & 
    % \vtop{
    %     \hbox{\strut 1. throat\_color\_grey}
    %     \hbox{\strut 2. \textbf{wing\_shape\_roundedwings}}
    %     \hbox{\strut 3. back\_color\_grey}
    % }
    & 
    \vtop{
        \hbox{\strut 1. throat\_color\_grey}
        \hbox{\strut 2. \textbf{wing\_shape\_roundedwings}}
        \hbox{\strut 3. back\_color\_grey}
    }
    % &
    % \vtop{
    %     \hbox{\strut 1. \textbf{wing\_shape\_roundedwings}}
    %     \hbox{\strut 2. wing\_pattern\_solid}
    %     \hbox{\strut 3. breast\_color\_black}
    % }
    & 
    \vtop{
        \hbox{\strut 1. underparts\_color\_grey}
        \hbox{\strut 2. back\_color\_grey}
        \hbox{\strut 3. belly\_color\_black}
    } \\ 
\midrule
Winter Wren &\vtop{
        \hbox{\strut 1. \textbf{breast\_color\_brown}}
        \hbox{\strut 2. \textbf{forehead\_color\_brown}}
        \hbox{\strut 3. \textbf{under\_tail\_color\_brown}}
    }& 
    \vtop{
        \hbox{\strut 1. \textbf{breast\_color\_brown}}
        \hbox{\strut 2. underparts\_color\_brown}
        \hbox{\strut 3. \textbf{forehead\_color\_brown}}
    }
    % & 
    % \vtop{ 
    %     \hbox{\strut 1. \textbf{breast\_color\_brown}}
    %     \hbox{\strut 2. underparts\_color\_brown}
    %     \hbox{\strut 3. \textbf{forehead\_color\_brown}}
    % }
    & 
    \vtop{
        \hbox{\strut 1. \textbf{breast\_color\_brown}}
        \hbox{\strut 2. underparts\_color\_brown}
        \hbox{\strut 3. \textbf{under\_tail\_color\_brown}}
    }
    % &
    % \vtop{
    %     \hbox{\strut 1. size\_very\_small\_3\_5\_in}
    %     \hbox{\strut 2. \textbf{breast\_color\_brown}}
    %     \hbox{\strut 3. nape\_color\_brown}
    % }
    & 
    \vtop{
        \hbox{\strut 1. size\_very\_small\_3\_5\_in}
        \hbox{\strut 2. \textbf{forehead\_color\_brown}}
        \hbox{\strut 3. crown\_color\_brown}
    } \\ 
\midrule
House Sparrow &\vtop{
        \hbox{\strut 1. \textbf{breast\_color\_grey}}
        \hbox{\strut 2. \textbf{upperparts\_color\_black}}
        \hbox{\strut 3. \textbf{primary\_color\_brown}}
    }&  
    \vtop{
        \hbox{\strut 1. \textbf{breast\_color\_grey}}
        \hbox{\strut 2. \textbf{primary\_color\_brown}}
        \hbox{\strut 3. bill\_shape\_cone}
    }
    % & 
    % \vtop{ 
    %     \hbox{\strut 1. \textbf{breast\_color\_grey}}
    %     \hbox{\strut 2. bill\_shape\_cone}
    %     \hbox{\strut 3. back\_color\_brown}
    % }
    & 
    \vtop{
        \hbox{\strut 1. bill\_shape\_cone}
        \hbox{\strut 2. \textbf{breast\_color\_grey}}
        \hbox{\strut 3. nape\_color\_grey}
    }
    % &
    % \vtop{
    %     \hbox{\strut 1. underparts\_color\_grey}
    %     \hbox{\strut 2. \textbf{breast\_color\_grey}}
    %     \hbox{\strut 3. upper\_tail\_color\_brown}
    % }
    & 
    \vtop{
        \hbox{\strut 1. \textbf{breast\_color\_grey}}
        \hbox{\strut 2. tail\_shape\_notched\_tail}
        \hbox{\strut 3. bill\_shape\_cone}
    } \\ 
\midrule
Prothonotary Warbler &\vtop{
        \hbox{\strut 1. \textbf{under\_tail\_color\_grey}}
        \hbox{\strut 2. \textbf{head\_pattern\_plain}}
        \hbox{\strut 3. \textbf{back\_pattern\_multicolored}}
    }&  
    \vtop{
        \hbox{\strut 1. head\_pattern\_plain}
        \hbox{\strut 2. leg\_color\_grey}
        \hbox{\strut 3. \textbf{back\_pattern\_multicolored}}
    }
    % & 
    % \vtop{
    %     \hbox{\strut 1. \textbf{back\_pattern\_multicolored}}
    %     \hbox{\strut 2. upper\_tail\_color\_grey}
    %     \hbox{\strut 3. \textbf{under\_tail\_color\_grey}}
    % }
    & 
    \vtop{
        \hbox{\strut 1. \textbf{back\_pattern\_multicolored}}
        \hbox{\strut 2. \textbf{under\_tail\_color\_grey}}
        \hbox{\strut 3. crown\_color\_yellow}
    }
    % &
    % \vtop{
    %     \hbox{\strut 1. crown\_color\_yellow}
    %     \hbox{\strut 2. \textbf{back\_pattern\_multicolored}}
    %     \hbox{\strut 3. \textbf{under\_tail\_color\_grey}}
    % }
    & 
    \vtop{
        \hbox{\strut 1. nape\_color\_yellow}
        \hbox{\strut 2. \textbf{back\_pattern\_multicolored}}
        \hbox{\strut 3. crown\_color\_yellow}
    } \\ 
\bottomrule
\end{tabular}
% \end{sc}
% \end{small}
\end{center}
% \vskip -0.1in
\end{table*}

\subsection{Retrieving the concepts from embeddings}
\noindent\textbf{Learn the concept predictor.}
For datasets with the concept annotation $\mathcal{C} \in \mathbb{R}^{N_c}$ ($N_c$ being the number of concepts per image $\mathcal{X}$), the learnable $t: R^l \rightarrow\mathcal{C}$ classifies the concepts using the embeddings $\Phi$ of each pruned networks. Per this definition, $t$ outputs a scalar value $c$ representing each concept in the concept vector for each input image. 
For datasets without concept annotation, we leverage a set of image embeddings with the concept being present and absent)~\cite{yuksekgonul2022post} and learn a linear SVM ($t$) to construct the concept activation matrix~\cite{kim2017interpretability} as $\boldsymbol{Q} \in\mathbb{R}^{N_c \times l}$. 
Finally we estimate the concept value as $c = \frac{<\Phi(x), q^i>}{||q_i||_2^2}$ $ \in \mathbb{R}$ utilizing each row $\boldsymbol{q^i}$ of $\boldsymbol{Q}$. Thus, the complete tuple of $j^{th}$ sample is $\{x_j, y_j, c_j\}$, denoting the image, label, and learned concept vector, respectively.

\noindent\textbf{Learn PCBM.}
Following~\cite{yuksekgonul2022post}, we carve the interpretable classifier $g: \mathcal{C} \rightarrow \mathcal{Y}$ from $f$ to predict the labels from the concepts using the following loss:
\begin{align}
\label{equ:unconstrained_risk}
\min_g \mathbb{E}_{c, y\sim\mathcal{C}, \mathcal{Y}} \big[\mathcal{L}(g(c), y) \big] + \frac{\lambda}{N_cK}\Omega(g),
\end{align}

where $g(c) = w^Tc + b$, $\lambda$ is the regularization strength, $\Omega$ is the elastic net penalty~\cite{yuksekgonul2022post}. $g$ associates each concept with a weight after training, implying its predictive significance to the class labels.

\subsection{Local explanations using Grad-CAM}
Saliency maps are heatmap-based techniques, highlighting essential features (pixels for images) in the input space responsible for the model's prediction as a class label $k$. In this paper, we adopt GRAD-CAM method~\cite{selvaraju2016grad}. We calculate the heatmap by choosing an intermediate convolutional layer and then linearizing the rest of the network to be interpretable.

\section{Experiments}
\begin{table*}[t]
\caption{The top-3 concepts with the highest weights of the carved interpretable models ($g$) using PCBM from the pruned subnetworks for various pruning iterations for the HAM10000 dataset. Due to the space constraint, we only report the concepts from $g$ extracted from the model for iterations 1, 2, 4, and 15 with 100\%, 90\%, 81\%, 72.9\%, 25.5\%, and 23\% weights remaining.
}
\fontsize{6.2pt}{0.20cm}\selectfont
\label{tab:ham10k}
% \vskip 0.15in
\begin{center}
% \begin{small}
% \begin{sc}
\begin{tabular}{l l l l l  }
\toprule 
        \textbf{LABEL} 
       & \textbf{100.0 \% WEIGHTS REMAINING} 
       & \textbf{90.0 \% WEIGHTS REMAINING} 
       % & \textbf{81.0 \% WEIGHTS REMAINING} 
       & \textbf{72.9 \% WEIGHTS REMAINING} 
       % & \textbf{25.5 \% WEIGHTS REMAINING} 
       & \textbf{23.0 \% WEIGHTS REMAINING} \\
\midrule 
    Malignant & 
    \vtop{
        \hbox{\strut 1. \textbf{BWV}}
        \hbox{\strut 2. \textbf{IrregularStreaks}}
        \hbox{\strut 3. \textbf{RegressionStructures}}
    }& 
    \vtop{
        \hbox{\strut 1. AtypicalPigmentNetwork}
        \hbox{\strut 2. \textbf{BWV}}
        \hbox{\strut 3. IrregularDG}
    }
    % & 
    % \vtop{
    %     \hbox{\strut 1. \textbf{IrregularStreaks}}
    %     \hbox{\strut 2. TypicalPigmentNetwork}
    %     \hbox{\strut 3. \textbf{BWV}}
    % }
    & 
    \vtop{
        \hbox{\strut 1. \textbf{IrregularStreaks}}
        \hbox{\strut 2. lAtypicalPigmentNetwork}
        \hbox{\strut 3. \textbf{RegressionStructures}}
    }
    % & 
    % \vtop{
    %     \hbox{\strut 1. \textbf{BWV}}
    %     \hbox{\strut 2. RegularStreaks}
    %     \hbox{\strut 3. \textbf{RegressionStructures}}
    % }
    & 
    \vtop{ 
        \hbox{\strut 1. AtypicalPigmentNetwork}
        \hbox{\strut 2. TypicalPigmentNetwork}
        \hbox{\strut 3. \textbf{IrregularStreaks}}
    }\\  
\midrule 
    Benign & 
    \vtop{
        \hbox{\strut 1. \textbf{TypicalPigmentNetwork}}
        \hbox{\strut 2. \textbf{RegularStreaks}}
        \hbox{\strut 3. \textbf{RegularDG}}
    }& 
    \vtop{
        \hbox{\strut 1. \textbf{TypicalPigmentNetwork}}
        \hbox{\strut 2. RegressionStructures}
        \hbox{\strut 3. \textbf{RegularDG}}
    }
    % & 
    % \vtop{
    %     \hbox{\strut 1. \textbf{RegularStreaks}}
    %     \hbox{\strut 2. \textbf{RegularDG}}
    %     \hbox{\strut 3. RegressionStructures}
    % }
    & 
    \vtop{
        \hbox{\strut 1. \textbf{RegularStreaks}}
        \hbox{\strut 2. \textbf{TypicalPigmentNetwork}}
        \hbox{\strut 3. \textbf{RegularDG}}
    }
    % & 
    % \vtop{
    %     \hbox{\strut 1. \textbf{RegularDG}}
    %     \hbox{\strut 2. RegressionStructures}
    %     \hbox{\strut 3. \textbf{RegularStreaks}}
    % }
    & 
    \vtop{
        \hbox{\strut 1. \textbf{RegularStreaks}}
        \hbox{\strut 2. \textbf{RegularDG}}
        \hbox{\strut 3. IrregularDG}
    }\\  
\bottomrule
\end{tabular}
\end{center}
\end{table*}

We perform experiments on CUB-200~\cite{wah2011caltech} and HAM10000~\cite{tschandl2018ham10000} using ResNet~\cite{he2016deep} and Inception~\cite{szegedy2015going} networks, respectively. We prune the networks for 15 iterations, removing 10\% of weights in each iteration, fine-tuning, and then pruning again. 
We conduct three experiments. First, we estimate the accuracy scores to evaluate the predictive performance of the different pruned networks and carved interpretable models. Second, we use interpretable models to estimate the top three concepts. Third, we compute the Grad-CAM-based saliency maps for each pruned network to compare the local explanations qualitatively. In all the subsequent plots, we denote the original network as the one with ``100\% weight remaining''. We use the entire $\Phi$ as the concept extractor. We describe the dataset and training configurations in detail in the supplementary materials. The code is available at: \url{https://github.com/batmanlab/lth-explain}

% We find that pruning and fine-tuning the model with LTH forces the network to concentrate on the relevant concepts, which improves the performance of the interpretable model $g$ relative to its pruned counterpart $f$. In addition, drastic pruning significantly modifies the pruned model. Therefore, it relies on irrelevant concepts, resulting in a performance decline.

\subsection{Results} 
\begin{figure}
\centering
\includegraphics[width=1\linewidth]{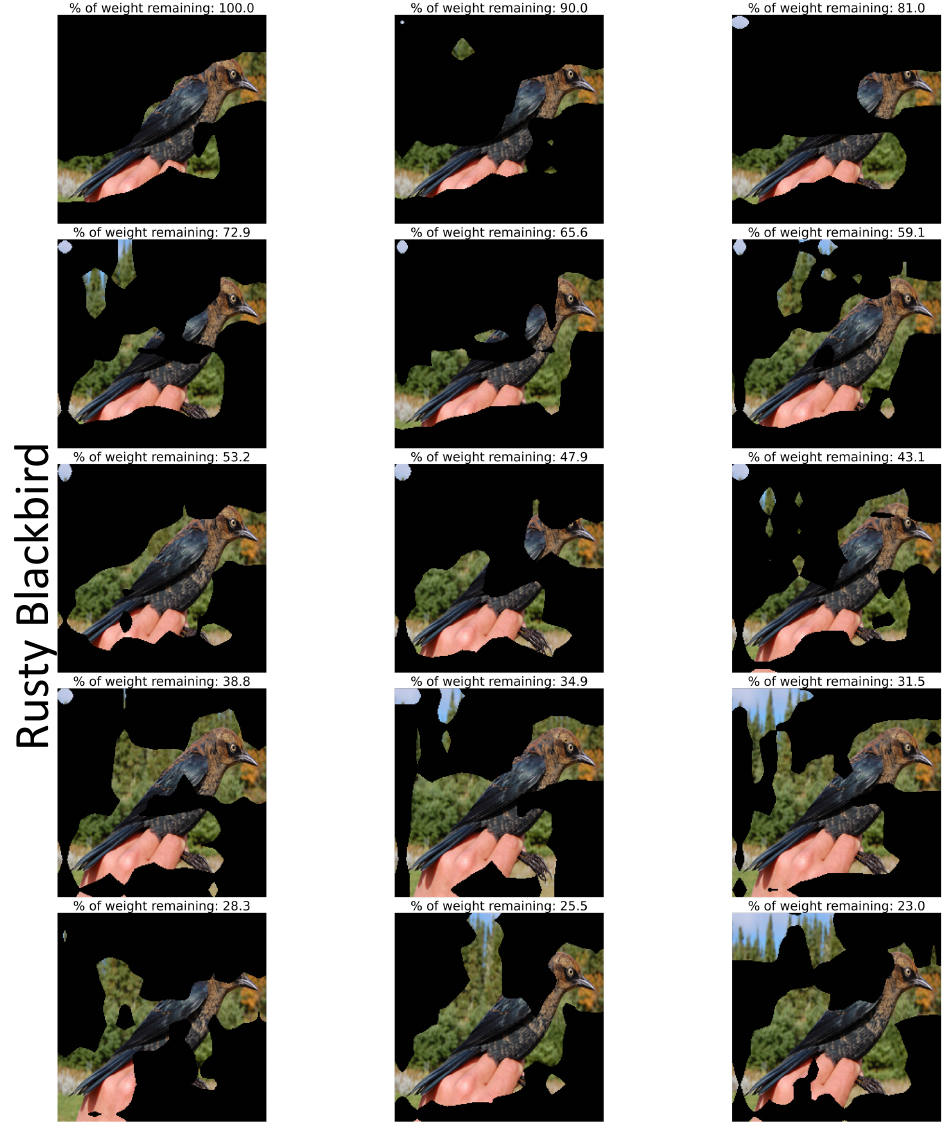}
\caption{Grad-CAM outputs of ``Rusty Blackbird'' for various pruned models.}
\label{fig:grad_cam_cub}
\end{figure}

\noindent\textbf{Quantitative evaluation of $\boldsymbol{f}$ and $\boldsymbol{g}$.}~\cref{fig:quant_accuracy} shows the accuracies for different pruning iterations for CUB-200 and HAM10000 datasets for the original neural network ($f$) and the carved interpretable model ($g$) using PCBM. As we prune more, the performance of $f$ and $g$ deteriorates. When $\sim$28\% weights remain, the accuracy of the $f$ and $g$ drops from 84.8\% and 75.7\% to 48\% and 66\% compared to the original networks for CUB-200. Strikingly, for later iterations, $g$ performs better than $f$. As pruning removes irrelevant components from the network, the pruned networks utilize the concepts for the classification. 

\noindent\textbf{Explaining LTH.}~\cref{tab:cub} and~\ref{tab:ham10k} reports the top-3 concepts for the different interpretable models $g$ for each pruned networks from $f$ for CUB-200 and HAM10000, respectively. After training, $g$ associates each concept with a weight. A concept with a high weight implies its high predictive significance.~\cref{tab:cub} and~\ref{tab:ham10k} demonstrate that different pruned networks rely on different concepts for classifying the same class labels. For example, the initial network with 100\% weights relies on \emph{breast\_color\_grey}, \emph{upperparts\_color\_black} and \emph{primary\_color\_brown} as top-3 concepts for the bird species ``House Sparrow''. However, when 23\% of the weights remained in the network, the classification of ``House Sparrow'' relies on \emph{tail\_shape\_notched\_tail} and \emph{bill\_shape\_cone} in addition to \emph{breast\_color\_grey}. For HAM10000, the initial networks with 100\% weights identifies \emph{Blue Whitish Veil (BWV)}, \emph{IrregularStreaks} and \emph{RegressionStructures} as top-3 concepts for Malignant skin lesion. These concepts are clinically relevant for malignancy~\cite{menzies1996sensitivity, lucieri2020interpretability}. However, the network with 23\% weights identifies clinically irrelevant \emph{TypicalPignmentNetwork} as one of the identifying concepts for malignancy. ~\cref{fig:grad_cam_cub} shows similar inconsistencies in the Grad-CAM outputs for different networks. The network with 100\% weights focuses on the relevant pixels on the bird's body, but the network with $\sim$ 23\% weights also highlights the background as relevant pixels.  These findings demonstrate that drastic pruning alters the network's representation, resulting in a performance decrease.
For more results, refer to the supplementary materials.

% \subsubsection{Evaluating performance of the pruned subnetworks}
% \input{Sections/results_quant.tex}

% \subsubsection{Evaluating global explanations of the pruned subnetworks}
% \input{Sections/results_TCAV.tex}

% \subsubsection{Evaluating local explanations of the pruned subnetworks}
% \input{Sections/results_Grad_CAM.tex}

% \input{Sections/results_cub_200.tex}

\section{Conclusion \& Future Work} 
In this paper, we study the success/failure modes of LTH using explainability with concepts and pixels. We observe that the pruned networks using LTH with more weights highlight relevant concepts and pixels; the networks with fewer weights do not. As a result, we conclude that magnitude iterative pruning does not emphasize the relevant concepts or pixels as the original model; in fact, the opposite is true. In the future, we want to extend this study to more real-life chest-x-ray datasets, eg. MIMIC-CXR~\cite{johnson2019mimic}. Also, we want to employ \emph{Route, interpret and repeat}~\cite{ghosh2023route} algorithm to rank the samples based on the \emph{difficulty} and investigate whether the extracted concepts differ significantly for "harder" samples as we prune. Also, a teacher-student framework can be employed where the original network and the subsequent pruned networks will be considered as teacher and student respectively. The saliency maps of the teacher model will ensure the student model focuses on the relevant pixels even with the limited network capacity.

%%%%%%%%% REFERENCES
{\small
\bibliographystyle{ieee_fullname}
\bibliography{egbib}
}

\section* {Supplementary materials}
\section{Reproduciblity}
The code is available at: \url{https://github.com/batmanlab/lth-explain}.
\section{Related Work}
\paragraph{Network pruning}
In practice, neural networks are frequently overparameterized. 
Distillation~\cite{ba2014deep, hinton2015distilling} and pruning~\cite{lecun1989optimal, han2015learning} both rely on the ability to reduce parameters while maintaining accuracy. 
Even with adequate memory capacity for training data, networks naturally learn more specific functions~\cite{zhang2021understanding, neyshabur2014search, arpit2017closer}
Later research shows that the overparameterized networks are easier to train~\cite{bengio2005convex, hinton2015distilling, zhou2016learning}. Recently Lottery Ticket Hypothesis~\cite{frankle2018lottery} aims to find a subnetwork within a deep neural net performs similarly or even better as the original deep network.

\paragraph{Saliency map based explanation methods}
First~\cite{simonyan2013deep} developed a saliency map technique to highlight the relevant pixels in the image for the model's prediction. In CAM~\cite{zhou2016learning}, we take the global average pooling of the feature map from the final convolution layer. Then we train a linear classifier to get the weights corresponding to each feature map, denoting the importance of a feature map to the final prediction. Later in~\cite{selvaraju2016grad, chattopadhyay2017grad, smilkov2017smoothgrad, sundararajan2017axiomatic} amalgamate the gradient information with relevant weights of the necessary pixels to generate more localized saliency maps. In summary, the saliency maps aim to provide local explanations in terms of pixels. However, these saliency maps are often criticized for being inconsistent ~\cite{pillai2022consistent}.~\cite{adebayo2018sanity, kindermans2017reliability} demonstrates that the saliency maps highlight the correct regions in the image even though the backbone's representation was arbitrarily perturbed. The explanations in terms of pixel intensities do not correspond to the high-level interpretable \emph{concept}, understood by humans. In this project, we also aim to provide the post hoc explanation of the all the pruned models in terms of the interpretable concepts as a global explanation, rather than the pixel intensities.

\paragraph{Concept-based explanation methods}
In \emph{concept} based explanation methods, researchers aim to quantify the importance of a humanly interpretable \emph{concept} for the model's prediction. In TCAV \cite{kim2017interpretability}, the researchers first learn the concept activation vectors (CAV) by learning a linear classifier that separates the concept images and the random images. After that, they estimate the TCAV score by taking the derivative of the prediction probabilities \wrt the concept activation vectors. In the concept completeness paper~\cite{yeh2019concept}, they derived a metric known as the ``concept completeness score'', which shows whether the given concepts are sufficient to explain the prediction of the black box. In the Network, dissection method \cite{bau2017network},  the activations of each unit are segmented to find its association with a given concept by taking the intersection over the union metric. Recently, in~\cite{crabbe2022concept}, the researchers relax the assumption in~\cite{kim2017interpretability} that concepts are linearly separable in the latent space; instead, they are represented by clusters, and the researchers obtain the concepts by utilizing the gaussian kernels for SVM based classifier. Also~\cite{koh2020concept} proposes the interpretable-by-design approcach ``Concept Bottleneck Models (CBMs)'', which predicts the concepts from the images and then the labels from the predicted concepts. Several modifications to the original CBM model are discussed~\cite{sarkar2021inducing,zarlenga2022concept,marconato2022glancenets}. Later~\cite{yuksekgonul2022post} proposes Post-hoc Concept Bottleneck Models (PCBMs), where they identify the concepts from the embeddings of a pre-trained model. Recently~\cite{ghosh2023route} introduces~\emph{Route, Interpret, and Repeat} where they carve out a mixture of interpretable models from a pre-trained neural network.

In this short paper, we want to study the explainability in terms of both the human interpretable concepts and pixels for the pruned subnetworks using LTH using PCBM and Grad-CAM, respectively.

\section{Background on LTH}
\emph{LTH}~\cite{frankle2018lottery} aims to find a subnetwork within a deep neural network that achieves similar accuracy as the original network -- \emph{when trained in isolation}. Specifically, we perform the following steps to train and prune the smallest magnitude weights of the neural network $f$:
\begin{enumerate}
\setlength{\itemsep}{1pt}
  \setlength{\parskip}{0pt}
  \setlength{\parsep}{0pt}
  \item Randomly initialize $f(.; \theta_0)$ (where $\theta_0 \sim \mathcal{D_\theta}$).
  \item Prune $p\%$ of the parameters $\theta_j$ with smallest magnitude weights, creating a mask $m$.
  \item Optimize the network for $j$ iterations using stochastic gradient descent (SGD) on a training set, arriving at parameters $\theta_j$.
  \item Reset the remaining parameters to their values in $\theta_0$, creating a subnetwork $f(x; 
m\odot\theta_0)$.
    \item Finetune and continue pruning for $n$ rounds obtaining $n$-pruned subnetworks.
\end{enumerate}
As pruning gradually reduces the network in size, we measure the performance and explainability for the all $n$-pruned subnetworks.

\section{Background on Grad-CAM}
Saliency maps are heatmap-based techniques, highlighting important features (pixels for images) in the input space responsible for the model's prediction as a class label $k$. In this project, we utilize GRAD-CAM method~\cite{selvaraju2016grad}. We calculate the heatmap by choosing an intermediate convolutional layer and then linearizing the rest of the network to be interpretable. Specifically, we estimate the derivative of the predicted output \wrt each channel of the convolutional layer averaged over all spatial locations as follows:
\begin{equation}
  w^m_k = \sum_i\sum_j \frac{\partial Y^k}{\partial A_{i,j}^m} \text{,} 
\end{equation} 
where $ w^k_c$ is the weight for $k^{th}$ class and $m^{th}$ feature map, $Y^k$ is the prediction for $k^{th}$ class and $A_{i,j}^m$ is the ${i, j}^{th}$ location of $m^{th}$ feature map. This results in a scalar for each channel that captures the importance of that channel in making the current prediction. Then, we calculate a weighted average of all activations of
the convolutional layer with the above-importance weights
for each channel to get a 2D matrix over spatial locations
Finally, we keep only positive numbers and resize them to the size of an input image to get the interpretation heatmap $e$. In this project, corresponding to an image, we estimate this heatmap for all the $n$-pruned networks.

\section{Dataset}
\noindent\textbf{CUB-200.}
The Caltech-UCSD Birds-200-2011~\cite{wah2011caltech} is a fine-grained classification dataset comprising 11788 images and 312 noisy visual concepts to classify the correct bird species from 200 possible classes. We adopted the strategy~\cite{barbiero2022entropy} to extract 108 denoised visual concepts. Also, we utilize training/validation splits from \cite{barbiero2022entropy}.

\noindent\textbf{HAM10000}
HAM10000 (\cite{tschandl2018ham10000}) is a classification dataset aiming to classify a skin lesion as benign or malignant. We follow the strategy in \cite{lucieri2020interpretability} to extract the 8 concepts from the Derm7pt (\cite{kawahara2018seven}) dataset.

\section{Training configurations}
\noindent\textbf{Configurations of LTH}
We utilize ResNet50~\cite{he2016deep} and Inception-V3~\cite{szegedy2015going} as the main neural network model to be pruned for CUB-200 and HAM10000 datasets, respectively. In each pruning iteration, we prune 10\% of the weights, and we prune it for 15 iterations. We resize the images to 448$\times$448 and 299$\times$299 for CUB-200 and HAM-10000, respectively. We set a batch size of 32 and utilized Stochastic gradient descent with a learning rate of 0.01 to train the main network and finetune the pruned subnetworks.

\noindent\textbf{Configurations of concept extractor}
We use all the convolution blocks of ResNet-50 and Inception-V3 as image embeddings. We flatten the image embeddings using adaptive average pooling to train the concept extractor $t$. For CUB-200, we train a SGD classifier for 15 epochs with a learning rate of 0.01. Lastly for HAM-10000, we train a linear SVM classifier to extract the concepts from the Derm7pt dataset. We use 50 images each of samples with and without the disease to train the SVM classifier.

\noindent\textbf{Training PCBM} The interpretable model $g$ of PCBM is an SGD classifier, trained for 35 epochs with a learning rate of 0.01.

\begin{table*}[t]
\caption{The top-3 concepts with the highest weights of the carved interpretable models ($g$) using PCBM from the pruned subnetworks for various pruning iterations for the bird species of the CUB-200 dataset. Due to the space constraint, we only report the concepts from $g$ extracted from the model for iterations 1, 2, 4, and 15 with 100\%, 90\%, 72.9\%, and 23\% weights remaining.}
\fontsize{6.2pt}{0.1cm}\selectfont
\label{tab:cub}
\begin{center}
\begin{tabular}{l l l l l }
\toprule 
        \textbf{LABEL} 
       & \textbf{100.0 \% WEIGHTS REMAINING} 
       & \textbf{90.0 \% WEIGHTS REMAINING} 
       % & \textbf{81.0 \% WEIGHTS REMAINING} 
       & \textbf{72.9 \% WEIGHTS REMAINING} 
       % & \textbf{25.5 \% WEIGHTS REMAINING} 
       & \textbf{23.0 \% WEIGHTS REMAINING} \\
\midrule 
Cerulean Warbler &\vtop{ 
        \hbox{\strut 1. \textbf{forehead\_color\_blue}}
        \hbox{\strut 2. \textbf{crown\_color\_blue}}
        \hbox{\strut 3. \textbf{nape\_color\_blue}}
    }&  
    \vtop{
        \hbox{\strut 1. \textbf{forehead\_color\_blue}}
        \hbox{\strut 2. \textbf{nape\_color\_blue}}
        \hbox{\strut 3. \textbf{crown\_color\_blue}}
    }
    & 
    \vtop{
        \hbox{\strut 1. \textbf{crown\_color\_blue}}
        \hbox{\strut 2. \textbf{nape\_color\_blue}}
        \hbox{\strut 3. size\_very\_small\_3\_5\_in}
    }
    & 
    \vtop{
        \hbox{\strut 1. size\_very\_small\_3\_5\_in}
        \hbox{\strut 2. \textbf{nape\_color\_blue}}
        \hbox{\strut 3. \textbf{forehead\_color\_blue}}
    } \\ 
\midrule
Rock Wren &\vtop{ 
        \hbox{\strut 1. \textbf{breast\_color\_buff}}
        \hbox{\strut 2. \textbf{belly\_color\_buff}}
        \hbox{\strut 3. \textbf{underparts\_color\_grey}}
    }& 
    \vtop{
        \hbox{\strut 1. \textbf{belly\_color\_buff}}
        \hbox{\strut 2. \textbf{breast\_color\_buff}}
        \hbox{\strut 3. upperparts\_color\_grey}
    }
    & 
    \vtop{ 
        \hbox{\strut 1. \textbf{belly\_color\_buff}}
        \hbox{\strut 2. \textbf{breast\_color\_buff}}
        \hbox{\strut 3. throat\_color\_grey}
    }
    & 
    \vtop{
        \hbox{\strut 1. breast\_color\_grey}
        \hbox{\strut 2. \textbf{belly\_color\_buff}}
        \hbox{\strut 3. leg\_color\_black}
    } \\ 
\midrule 
Loggerhead Shriker &\vtop{
        \hbox{\strut 1. \textbf{tail\_pattern\_multicolored}}
        \hbox{\strut 2. \textbf{forehead\_color\_grey}}
        \hbox{\strut 3. \textbf{crown\_color\_grey}}
    }&  
    \vtop{
        \hbox{\strut 1. \textbf{forehead\_color\_grey}}
        \hbox{\strut 2. \textbf{crown\_color\_grey}}
        \hbox{\strut 3. leg\_color\_black}
    }
    & 
    \vtop{
        \hbox{\strut 1.\textbf{crown\_color\_grey}}
        \hbox{\strut 2. \textbf{forehead\_color\_grey}}
        \hbox{\strut 3. head\_pattern\_capped}
    }
    & 
    \vtop{
        \hbox{\strut 1. \textbf{forehead\_color\_grey}}
        \hbox{\strut 2. leg\_color\_black}
        \hbox{\strut 3. primary\_color\_buff}
    } \\ 
\midrule 
Philadelphia Vireo &\vtop{
        \hbox{\strut 1. \textbf{size\_very\_small\_3\_5\_in}}
        \hbox{\strut 2. \textbf{under\_tail\_color\_grey}}
        \hbox{\strut 3. \textbf{leg\_color\_grey}}
    }&  
    \vtop{
        \hbox{\strut 1. \textbf{size\_very\_small\_3\_5\_in}}
        \hbox{\strut 2. \textbf{leg\_color\_grey}}
        \hbox{\strut 3. forehead\_color\_grey}
    }
    & 
    \vtop{
        \hbox{\strut 1. \textbf{primary\_color\_buff}}
        \hbox{\strut 2. \textbf{under\_tail\_color\_grey}}
        \hbox{\strut 3. \textbf{size\_very\_small\_3\_5\_in}}
    }
    & 
    \vtop{
        \hbox{\strut 1. crown\_color\_grey}
        \hbox{\strut 2.\textbf{leg\_color\_grey}}
        \hbox{\strut 3. \textbf{under\_tail\_color\_grey}}
    } \\ 
\midrule 
Tennessee Warbler &\vtop{
        \hbox{\strut 1. \textbf{bill\_color\_grey}}
        \hbox{\strut 2. \textbf{wing\_color\_yellow}}
        \hbox{\strut 3. \textbf{leg\_color\_grey}}
    }&  
    \vtop{
        \hbox{\strut 1. under\_tail\_color\_grey}
        \hbox{\strut 2. \textbf{bill\_color\_grey}}
        \hbox{\strut 3. \textbf{wing\_color\_yellow}}
    }
    % & 
    % \vtop{
    %     \hbox{\strut 1. \textbf{bill\_color\_grey}}
    %     \hbox{\strut 2. under\_tail\_color\_grey}
    %     \hbox{\strut 3. \textbf{wing\_color\_yellow}}
    % }
    & 
    \vtop{
        \hbox{\strut 1. \textbf{bill\_color\_grey}}
        \hbox{\strut 2. under\_tail\_color\_grey}
        \hbox{\strut 3. forehead\_color\_yellow}
    }
    % &
    % \vtop{
    %     \hbox{\strut 1. forehead\_color\_yellow}
    %     \hbox{\strut 2. \textbf{bill\_color\_grey}}
    %     \hbox{\strut 3. size\_very\_small\_3\_5\_in}
    % }
    & 
    \vtop{
        \hbox{\strut 1. \textbf{bill\_color\_grey}}
        \hbox{\strut 2. \textbf{leg\_color\_grey}}
        \hbox{\strut 3. forehead\_color\_yellow}
    } \\ 
    \midrule 
Palm Warbler &\vtop{
        \hbox{\strut 1. \textbf{wing\_pattern\_striped}}
        \hbox{\strut 2. \textbf{wing\_shape\_roundedwings}}
        \hbox{\strut 3. \textbf{leg\_color\_black}}
    }&  
    \vtop{
        \hbox{\strut 1. \textbf{wing\_pattern\_striped}}
        \hbox{\strut 2. throat\_color\_yellow}
        \hbox{\strut 3. bill\_shape\_allpurpose}
    }
    % & 
    % \vtop{
    %     \hbox{\strut 1. \textbf{wing\_pattern\_striped}}
    %     \hbox{\strut 2. throat\_color\_yellow}
    %     \hbox{\strut 3. bill\_shape\_allpurpose}
    % }
    & 
    \vtop{
        \hbox{\strut 1. \textbf{wing\_pattern\_striped}}
        \hbox{\strut 2. throat\_color\_yellow}
        \hbox{\strut 3. bill\_shape\_allpurpose}
    }
    % &
    % \vtop{
    %     \hbox{\strut 1. \textbf{wing\_pattern\_striped}}
    %     \hbox{\strut 2. \textbf{wing\_shape\_roundedwings}}
    %     \hbox{\strut 3. bill\_shape\_allpurpose}
    % }
    & 
    \vtop{
        \hbox{\strut 1. \textbf{wing\_pattern\_striped}}
        \hbox{\strut 2. \textbf{wing\_shape\_roundedwings}}
        \hbox{\strut 3. bill\_shape\_allpurpose}
    } \\ 
    \midrule 
Mourning Warbler &\vtop{
        \hbox{\strut 1. \textbf{leg\_color\_buff}}
        \hbox{\strut 2. \textbf{under\_tail\_color\_grey}}
        \hbox{\strut 3. \textbf{throat\_color\_grey}}
    }&  
    \vtop{
        \hbox{\strut 1. crown\_color\_black}
        \hbox{\strut 2. \textbf{throat\_color\_grey}}
        \hbox{\strut 3. forehead\_color\_black}
    }
    % & 
    % \vtop{
    %     \hbox{\strut 1. \textbf{throat\_color\_grey}}
    %     \hbox{\strut 2. crown\_color\_black}
    %     \hbox{\strut 3. nape\_color\_grey}
    % }
    & 
    \vtop{
        \hbox{\strut 1. under\_tail\_color\_yellow}
        \hbox{\strut 2. \textbf{leg\_color\_buff}}
        \hbox{\strut 3. crown\_color\_black}
    }
    % &
    % \vtop{
    %     \hbox{\strut 1. throat\_color\_grey}
    %     \hbox{\strut 2. wing\_pattern\_solid}
    %     \hbox{\strut 3. \textbf{leg\_color\_buff}}
    % }
    & 
    \vtop{
        \hbox{\strut 1. \textbf{leg\_color\_buff}}
        \hbox{\strut 2. wing\_color\_yellow}
        \hbox{\strut 3. wing\_pattern\_solid}
    } \\ 
    \midrule 
Wilson Warbler &\vtop{
        \hbox{\strut 1. \textbf{crown\_color\_black}}
        \hbox{\strut 2. \textbf{under\_tail\_color\_yellow}}
        \hbox{\strut 3. \textbf{head\_pattern\_capped}}
    }&  
    \vtop{
        \hbox{\strut 1. \textbf{crown\_color\_black}}
        \hbox{\strut 2. nape\_color\_yellow}
        \hbox{\strut 3. \textbf{under\_tail\_color\_yellow}}
    }
    % & 
    % \vtop{
    %     \hbox{\strut 1. \textbf{crown\_color\_black}}
    %     \hbox{\strut 2. back\_color\_yellow}
    %     \hbox{\strut 3. nape\_color\_yellow}
    % }
    & 
    \vtop{
        \hbox{\strut 1. \textbf{under\_tail\_color\_yellow}}
        \hbox{\strut 2. \textbf{crown\_color\_black}}
        \hbox{\strut 3. head\_pattern\_plain}
    }
    % &
    % \vtop{
    %     \hbox{\strut 1. back\_color\_yellow}
    %     \hbox{\strut 2. forehead\_color\_yellow}
    %     \hbox{\strut 3. head\_pattern\_plain}
    % }
    & 
    \vtop{
        \hbox{\strut 1. \textbf{under\_tail\_color\_yellow}}
        \hbox{\strut 2. size\_very\_small\_3\_5\_in}
        \hbox{\strut 3. back\_color\_yellow}
    } \\ 
    \midrule
Cactus Wren &\vtop{
        \hbox{\strut 1. \textbf{belly\_color\_buff}}
        \hbox{\strut 2. \textbf{breast\_color\_black}}
        \hbox{\strut 3. \textbf{underparts\_color\_buff}}
    }&  
    \vtop{
        \hbox{\strut 1. \textbf{breast\_color\_black}}
        \hbox{\strut 2. \textbf{underparts\_color\_buff}}
        \hbox{\strut 3. bill\_length\_about\_same\_as\_head}
    }
    % & 
    % \vtop{
    %     \hbox{\strut 1. \textbf{belly\_color\_buff}}
    %     \hbox{\strut 2. \textbf{breast\_color\_black}}
    %     \hbox{\strut 3. throat\_color\_white}
    % }
    & 
    \vtop{
        \hbox{\strut 1. \textbf{underparts\_color\_buff}}
        \hbox{\strut 2. bill\_length\_about\_same\_as\_head}
        \hbox{\strut 3. \textbf{breast\_color\_black}}
    }
    % &
    % \vtop{
    %     \hbox{\strut 1. throat\_color\_white}
    %     \hbox{\strut 2. wing\_pattern\_striped}
    %     \hbox{\strut 3. leg\_color\_black}
    % }
    & 
    \vtop{
        \hbox{\strut 1. bill\_shape\_dagger}
        \hbox{\strut 2. \textbf{belly\_color\_buff}}
        \hbox{\strut 3. leg\_color\_black}
    } \\ 
\midrule
Elegant Tern &\vtop{
        \hbox{\strut 1. \textbf{bill\_shape\_dagger}}
        \hbox{\strut 2. \textbf{head\_pattern\_capped}}
        \hbox{\strut 3. \textbf{wing\_color\_grey}}
    }& 
    \vtop{
        \hbox{\strut 1. \textbf{bill\_shape\_dagger}}
        \hbox{\strut 2. \textbf{head\_pattern\_capped}}
        \hbox{\strut 3. back\_color\_white}
    }
    & 
    \vtop{
        \hbox{\strut 1. \textbf{bill\_shape\_dagger}}
        \hbox{\strut 2. back\_color\_white}
        \hbox{\strut 3. \textbf{head\_pattern\_capped}}
    }
    & 
    \vtop{
        \hbox{\strut 1. \textbf{bill\_shape\_dagger}}
        \hbox{\strut 2. \textbf{head\_pattern\_capped}}
        \hbox{\strut 3. back\_color\_white}
    } \\ 
\midrule
Savannah Sparrow &\vtop{
        \hbox{\strut 1. \textbf{breast\_pattern\_striped}}
        \hbox{\strut 2. \textbf{tail\_shape\_notched\_tail}}
        \hbox{\strut 3. \textbf{throat\_color\_yellow}}
    }& 
    \vtop{
        \hbox{\strut 1. \textbf{breast\_pattern\_striped}}
        \hbox{\strut 2. \textbf{breast\_color\_black}}
        \hbox{\strut 3. crown\_color\_black}
    }
    & 
    \vtop{
        \hbox{\strut 1. \textbf{breast\_pattern\_striped}}
        \hbox{\strut 2. \textbf{throat\_color\_yellow}}
        \hbox{\strut 3. breast\_color\_black}
    }
    & 
    \vtop{
        \hbox{\strut 1. \textbf{breast\_pattern\_striped}}
        \hbox{\strut 2. breast\_color\_black}
        \hbox{\strut 3. under\_tail\_color\_buff}
    } \\ 
\midrule
Louisiana Waterthrush &\vtop{
        \hbox{\strut 1. \textbf{wing\_pattern\_solid}}
        \hbox{\strut 2. \textbf{back\_pattern\_solid}}
        \hbox{\strut 3. \textbf{underparts\_color\_white}}
    }& 
    \vtop{
        \hbox{\strut 1. breast\_pattern\_striped}
        \hbox{\strut 2. \textbf{wing\_pattern\_solid}}
        \hbox{\strut 3. throat\_color\_black}
    }
    & 
    \vtop{
        \hbox{\strut 1. leg\_color\_buff}
        \hbox{\strut 2. \textbf{wing\_pattern\_solid}}
        \hbox{\strut 3. breast\_pattern\_striped}
    }
    & 
    \vtop{
        \hbox{\strut 1. tail\_pattern\_solid}
        \hbox{\strut 2. belly\_color\_white}
        \hbox{\strut 3. breast\_color\_white}
    } \\ 
\midrule
Cape May Warbler &\vtop{
        \hbox{\strut 1. \textbf{breast\_pattern\_striped}}
        \hbox{\strut 2. \textbf{breast\_color\_black}}
        \hbox{\strut 3. \textbf{wing\_pattern\_striped}}
    }& 
    \vtop{
        \hbox{\strut 1. \textbf{breast\_pattern\_striped}}
        \hbox{\strut 2. \textbf{breast\_color\_black}}
        \hbox{\strut 3.  tail\_shape\_notched\_tail}
    }
    & 
    \vtop{
        \hbox{\strut 1. underparts\_color\_black}
        \hbox{\strut 2. tail\_shape\_notched\_tail}
        \hbox{\strut 3. \textbf{breast\_color\_black}}
    }
    & 
    \vtop{
        \hbox{\strut 1. \textbf{wing\_pattern\_striped}}
        \hbox{\strut 2. tail\_shape\_notched\_tail}
        \hbox{\strut 3. underparts\_color\_black}
    } \\
\midrule
Cliff Swallow &\vtop{ 
        \hbox{\strut 1. \textbf{belly\_color\_buff}}
        \hbox{\strut 2. \textbf{underparts\_color\_buff}}
        \hbox{\strut 3. \textbf{bill\_shape\_cone}}
    }& 
    \vtop{
        \hbox{\strut 1. \textbf{underparts\_color\_buff}}
        \hbox{\strut 2. \textbf{belly\_color\_buff}}
        \hbox{\strut 3. \textbf{bill\_shape\_cone}}
    }
    & 
    \vtop{
        \hbox{\strut 1. \textbf{belly\_color\_buff}}
        \hbox{\strut 2. \textbf{underparts\_color\_buff}}
        \hbox{\strut 3. \textbf{bill\_shape\_cone}}
    }
    & 
    \vtop{
        \hbox{\strut 1. \textbf{belly\_color\_buff}}
        \hbox{\strut 2. \textbf{underparts\_color\_buff}}
        \hbox{\strut 3. bill\_color\_black}
    } \\
\midrule
Caspian Tern &\vtop{ 
        \hbox{\strut 1. \textbf{leg\_color\_black}}
        \hbox{\strut 2. \textbf{tail\_pattern\_solid}}
        \hbox{\strut 3. \textbf{upper\_tail\_color\_white}}
    }& 
    \vtop{
        \hbox{\strut 1. \textbf{tail\_pattern\_solid}}
        \hbox{\strut 2. \textbf{leg\_color\_black}}
        \hbox{\strut 3. forehead\_color\_red}
    }
    & 
    \vtop{ 
        \hbox{\strut 1. \textbf{leg\_color\_black}}
        \hbox{\strut 2. \textbf{tail\_pattern\_solid}}
        \hbox{\strut 3. crown\_color\_black}
    }
    & 
    \vtop{
        \hbox{\strut 1. bill\_shape\_dagger}
        \hbox{\strut 2. head\_pattern\_capped}
        \hbox{\strut 3. \textbf{leg\_color\_black}}
    } \\
\midrule
Red eyed Vireo &\vtop{
        \hbox{\strut 1. \textbf{upper\_tail\_color\_grey}}
        \hbox{\strut 2. \textbf{forehead\_color\_yellow}}
        \hbox{\strut 3. \textbf{back\_pattern\_solid}}
    }& 
    \vtop{
        \hbox{\strut 1. throat\_color\_white}
        \hbox{\strut 2. \textbf{upper\_tail\_color\_grey}}
        \hbox{\strut 3. \textbf{forehead\_color\_yellow}}
    }
    & 
    \vtop{ 
        \hbox{\strut 1. bill\_color\_grey}
        \hbox{\strut 2. tail\_pattern\_multicolored}
        \hbox{\strut 3. forehead\_color\_grey}
    }
    & 
    \vtop{
        \hbox{\strut 1. \textbf{forehead\_color\_yellow}}
        \hbox{\strut 2. underparts\_color\_white}
        \hbox{\strut 3. breast\_pattern\_solid}
    } \\
\midrule
Chestnut sided Warbler &\vtop{
        \hbox{\strut 1. \textbf{crown\_color\_yellow}}
        \hbox{\strut 2. \textbf{forehead\_color\_yellow}}
        \hbox{\strut 3. \textbf{upper\_tail\_color\_grey}}
    }& 
    \vtop{
        \hbox{\strut 1. \textbf{forehead\_color\_yellow}}
        \hbox{\strut 2. \textbf{crown\_color\_yellow}}
        \hbox{\strut 3. \textbf{upper\_tail\_color\_grey}}
    }
    & 
    \vtop{
        \hbox{\strut 1. \textbf{forehead\_color\_yellow}}
        \hbox{\strut 2. \textbf{crown\_color\_yellow}}
        \hbox{\strut 3. back\_pattern\_striped}
    }
    & 
    \vtop{
        \hbox{\strut 1. \textbf{crown\_color\_yellow}}
        \hbox{\strut 2. wing\_color\_white}
        \hbox{\strut 3. tail\_shape\_notched\_tail'}
    } \\
\midrule
Green tailed Towhee &\vtop{ 
        \hbox{\strut 1. \textbf{belly\_color\_grey}}
        \hbox{\strut 2. \textbf{bill\_color\_grey}}
        \hbox{\strut 3. \textbf{wing\_color\_yellow}}
    }& 
    \vtop{
        \hbox{\strut 1. \textbf{belly\_color\_grey}}
        \hbox{\strut 2. underparts\_color\_grey}
        \hbox{\strut 3. \textbf{wing\_color\_yellow}}
    }
    & 
    \vtop{
        \hbox{\strut 1. underparts\_color\_grey}
        \hbox{\strut 2. \textbf{bill\_color\_grey}}
        \hbox{\strut 3. \textbf{belly\_color\_grey}}
    }
    & 
    \vtop{
        \hbox{\strut 1. underparts\_color\_grey}
        \hbox{\strut 2. \textbf{wing\_color\_yellow}}
        \hbox{\strut 3. \textbf{belly\_color\_grey}}
    } \\
\midrule
Kentucky Warbler &\vtop{
        \hbox{\strut 1. \textbf{leg\_color\_buff}}
        \hbox{\strut 2. \textbf{crown\_color\_black}}
        \hbox{\strut 3. \textbf{throat\_color\_yellow}}
    }
    & 
    \vtop{
        \hbox{\strut 1. \textbf{leg\_color\_buff}}
        \hbox{\strut 2. \textbf{crown\_color\_black}}
        \hbox{\strut 3. \textbf{throat\_color\_yellow}}
    }
    & 
    \vtop{
        \hbox{\strut 1. \textbf{leg\_color\_buff}}
        \hbox{\strut 2. forehead\_color\_black}
        \hbox{\strut 3. \textbf{crown\_color\_black}}
    } 
    & 
    \vtop{
        \hbox{\strut 1. wing\_color\_yellow}
        \hbox{\strut 2. \textbf{leg\_color\_buff}}
        \hbox{\strut 3. under\_tail\_color\_yellow}
    } \\
\midrule
Forsters Tern &\vtop{
        \hbox{\strut 1. \textbf{head\_pattern\_capped}}
        \hbox{\strut 2. \textbf{bill\_shape\_dagger}}
        \hbox{\strut 3. \textbf{nape\_color\_black}}
    }
    & 
    \vtop{
        \hbox{\strut 1. \textbf{head\_pattern\_capped}}
        \hbox{\strut 2. \textbf{bill\_shape\_dagger}}
        \hbox{\strut 3. forehead\_color\_black}
    }
    & 
    \vtop{
        \hbox{\strut 1. breast\_color\_grey}
        \hbox{\strut 2. \textbf{head\_pattern\_capped}}
        \hbox{\strut 3. crown\_color\_black}
    } 
    & 
    \vtop{
        \hbox{\strut 1. \textbf{head\_pattern\_capped}}
        \hbox{\strut 2. \textbf{bill\_shape\_dagger}}
        \hbox{\strut 3. crown\_color\_black}
    } \\
\midrule
Mourning Warbler &\vtop{
        \hbox{\strut 1. \textbf{leg\_color\_buff}}
        \hbox{\strut 2. \textbf{crown\_color\_black}}
        \hbox{\strut 3. \textbf{throat\_color\_grey}}
    }
    & 
    \vtop{
        \hbox{\strut 1. \textbf{crown\_color\_black}}
        \hbox{\strut 2. \textbf{throat\_color\_grey}}
        \hbox{\strut 3. forehead\_color\_black}
    }
    & 
    \vtop{
        \hbox{\strut 1. under\_tail\_color\_yellow}
        \hbox{\strut 2. \textbf{leg\_color\_buff}}
        \hbox{\strut 3. \textbf{crown\_color\_black}}
    } 
    & 
    \vtop{
        \hbox{\strut 1. \textbf{leg\_color\_buff}}
        \hbox{\strut 2. wing\_color\_yellow}
        \hbox{\strut 3. wing\_pattern\_solid}
    } \\
\midrule
Clark Nutcracker &\vtop{
        \hbox{\strut 1. \textbf{breast\_color\_grey}}
        \hbox{\strut 2. \textbf{upper\_tail\_color\_black}}
        \hbox{\strut 3. \textbf{primary\_color\_grey}}
    }
    & 
    \vtop{
        \hbox{\strut 1. \textbf{breast\_color\_grey}}
        \hbox{\strut 2. under\_tail\_color\_white}
        \hbox{\strut 3. wing\_pattern\_multicolored}
    }
    & 
    \vtop{
        \hbox{\strut 1. underparts\_color\_grey}
        \hbox{\strut 2. under\_tail\_color\_white}
        \hbox{\strut 3. \textbf{primary\_color\_grey}}
    } 
    & 
    \vtop{
        \hbox{\strut 1. \textbf{breast\_color\_grey}}
        \hbox{\strut 2. throat\_color\_grey}
        \hbox{\strut 3. belly\_color\_grey}
    } \\
\midrule
Hooded Oriole &\vtop{
        \hbox{\strut 1. \textbf{tail\_pattern\_multicolored}}
        \hbox{\strut 2. \textbf{throat\_color\_black}}
        \hbox{\strut 3. \textbf{wing\_pattern\_multicolored}}
    }
    & 
    \vtop{
        \hbox{\strut 1. leg\_color\_grey}
        \hbox{\strut 2. back\_color\_yellow}
        \hbox{\strut 3. \textbf{wing\_pattern\_multicolored}}
    }
    & 
    \vtop{
        \hbox{\strut 1. leg\_color\_grey}
        \hbox{\strut 2. wing\_color\_white}
        \hbox{\strut 3. \textbf{wing\_pattern\_multicolored}}
    } 
    & 
    \vtop{
        \hbox{\strut 1. leg\_color\_grey}
        \hbox{\strut 2. \textbf{wing\_pattern\_multicolored}}
        \hbox{\strut 3. head\_pattern\_plain}
    } \\
\midrule
Gray Catbird &\vtop{
        \hbox{\strut 1. \textbf{throat\_color\_grey}}
        \hbox{\strut 2. \textbf{head\_pattern\_capped}}
        \hbox{\strut 3. \textbf{upper\_tail\_color\_grey}}
    }
    & 
    \vtop{
        \hbox{\strut 1. \textbf{upper\_tail\_color\_grey}}
        \hbox{\strut 2. \textbf{throat\_color\_grey}}
        \hbox{\strut 3. back\_color\_grey}
    }
    & 
    \vtop{
        \hbox{\strut 1. \textbf{throat\_color\_grey}}
        \hbox{\strut 2. primary\_color\_grey}
        \hbox{\strut 3. under\_tail\_color\_grey}
    } 
    & 
    \vtop{
        \hbox{\strut 1. \textbf{throat\_color\_grey}}
        \hbox{\strut 2. \textbf{back\_color\_grey}}
        \hbox{\strut 3. wing\_pattern\_solid}
    } \\

\bottomrule
\end{tabular}
% \end{sc}
% \end{small}
\end{center}
% \vskip -0.1in
\end{table*}

\begin{figure*}[h]
\centering
\includegraphics[width=1\linewidth]{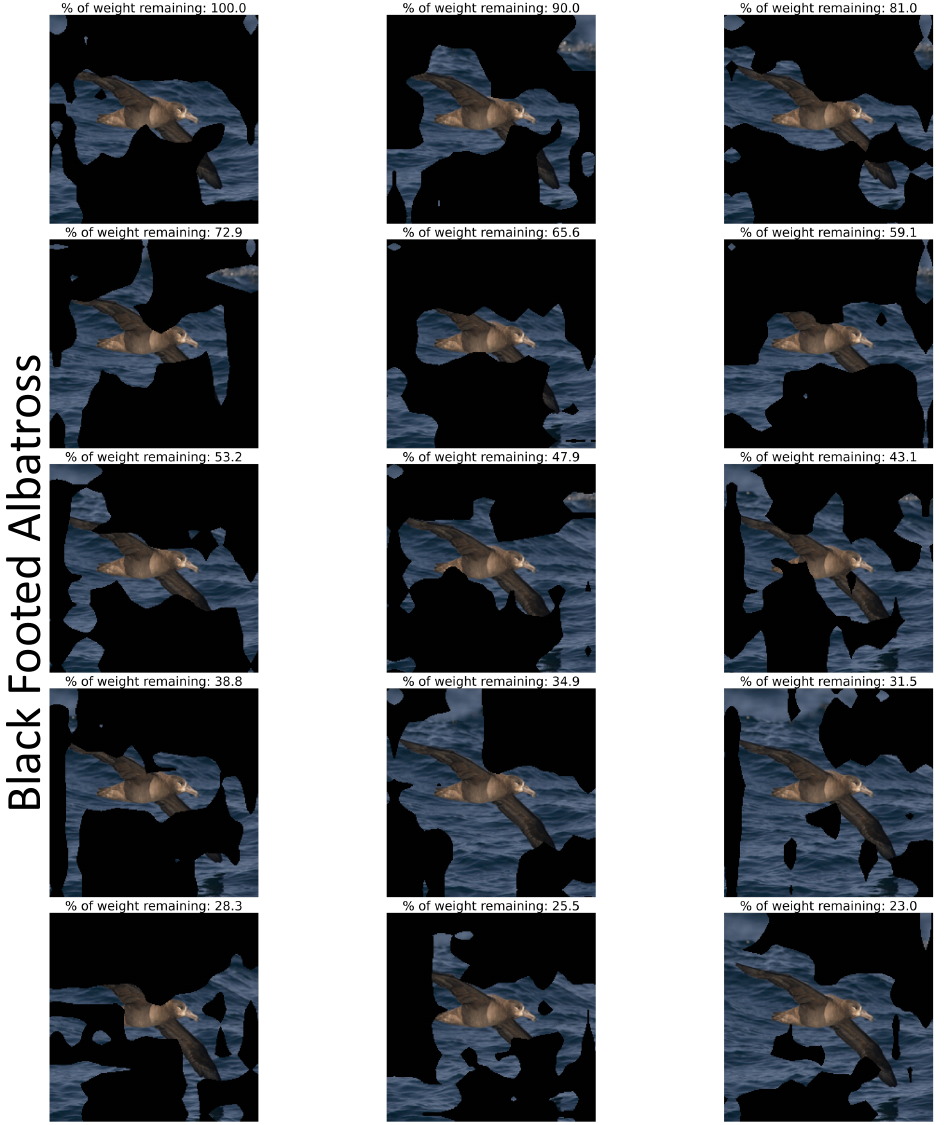}
\caption{Grad-CAM outputs of ``Black Footed Albatross'' for various pruned models.}
\label{fig:grad_cam_cub}
\end{figure*}

\begin{figure*}[h]
\centering
\includegraphics[width=1\linewidth]{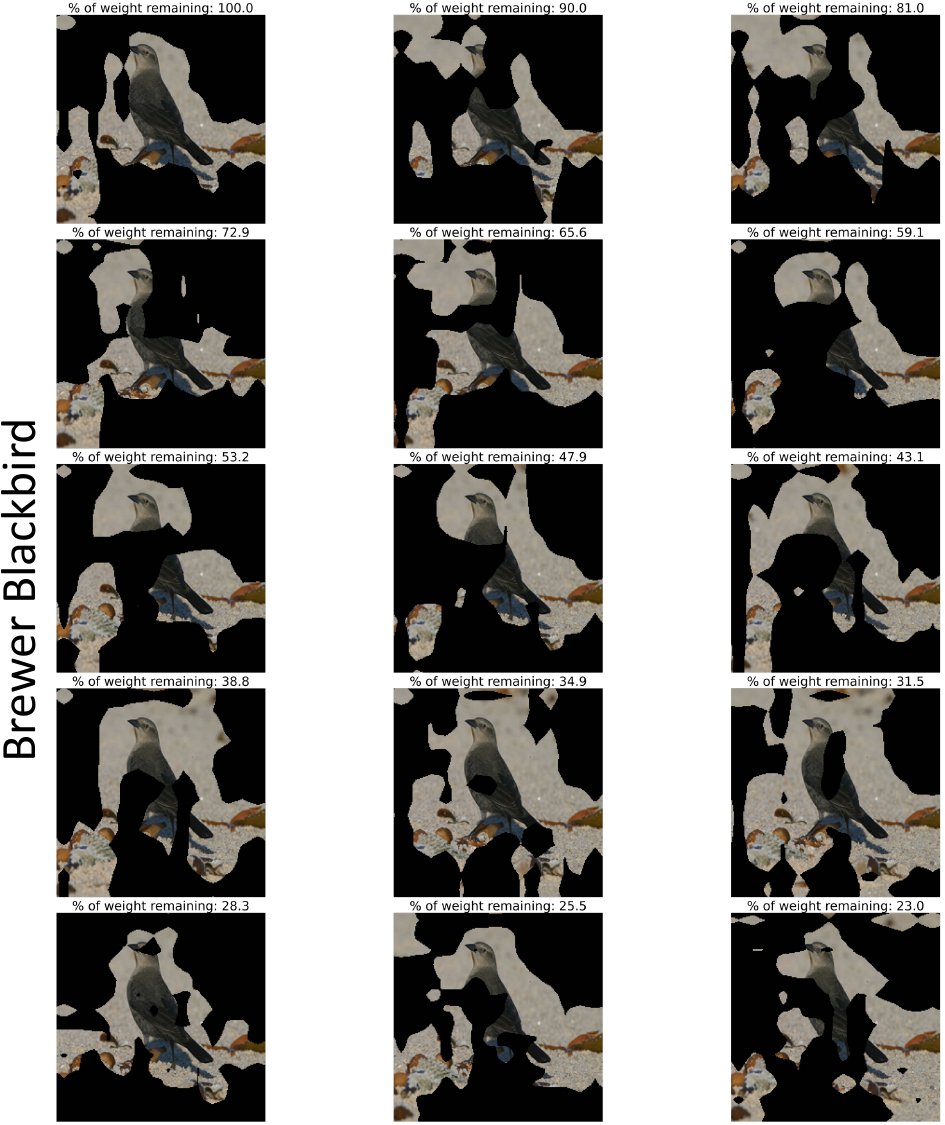}
\caption{Grad-CAM outputs of ``Brewer Blackbird'' for various pruned models.}
\label{fig:grad_cam_cub}
\end{figure*}

\begin{figure*}[h]
\centering
\includegraphics[width=1\linewidth]{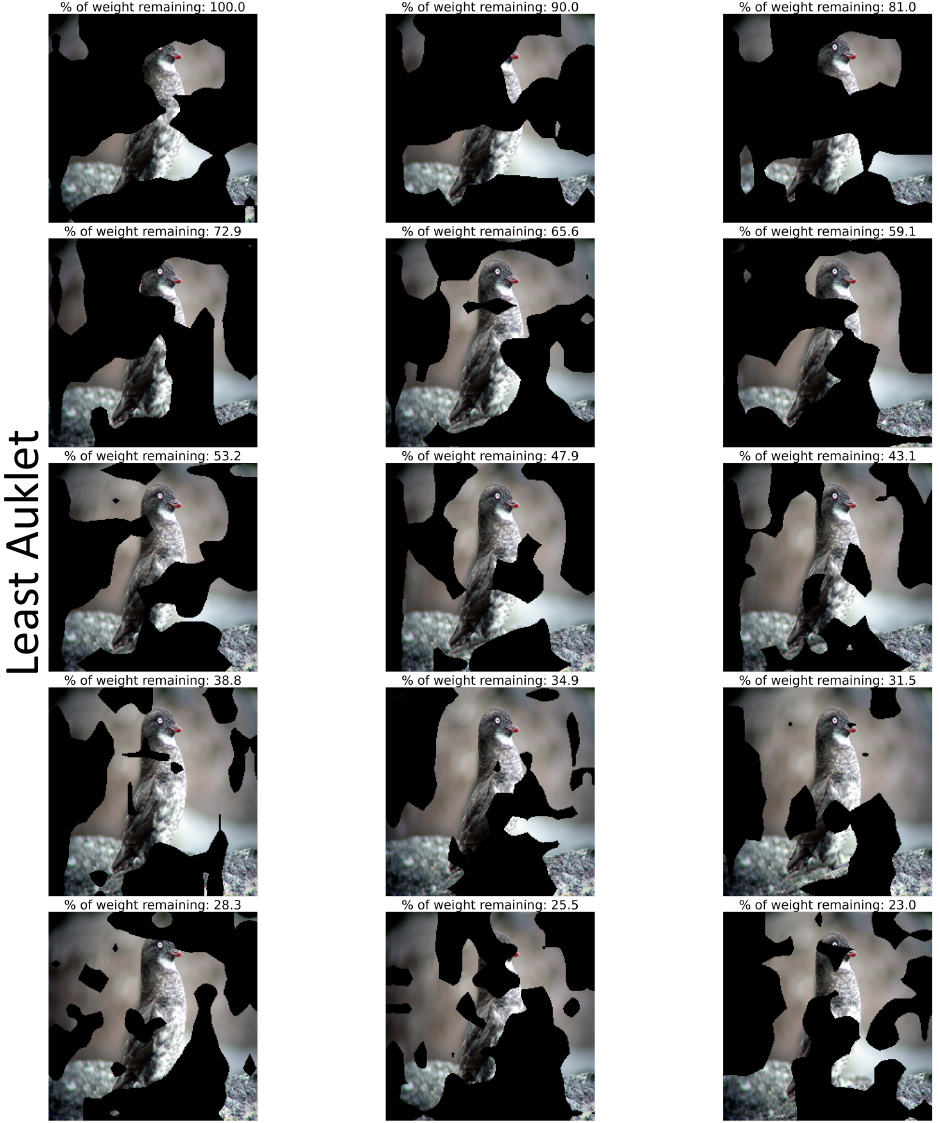}
\caption{Grad-CAM outputs of ``Least Auklet'' for various pruned models.}
\label{fig:grad_cam_cub}
\end{figure*}

\begin{figure*}[h]
\centering
\includegraphics[width=1\linewidth]{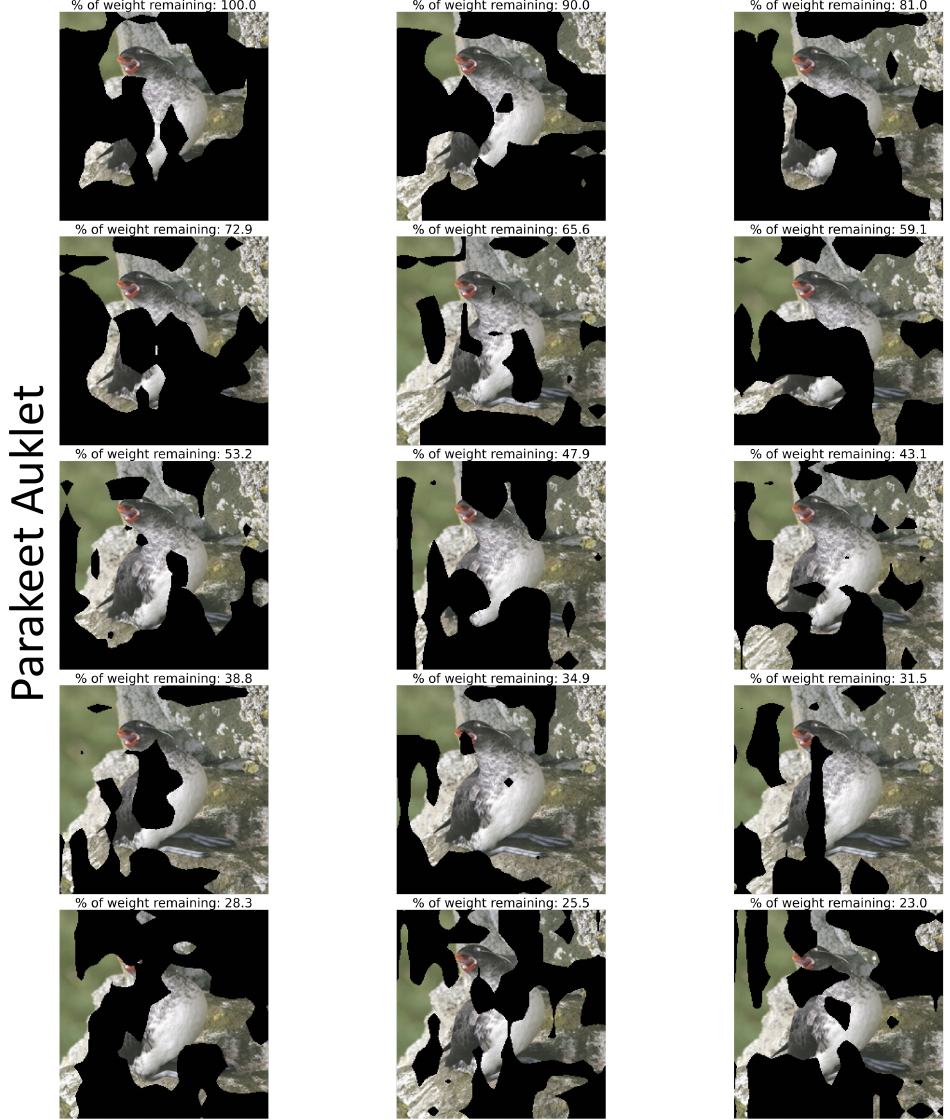}
\caption{Grad-CAM outputs of ``Parakeet Auklet'' for various pruned models.}
\label{fig:grad_cam_cub}
\end{figure*}

\begin{figure*}[h]
\centering
\includegraphics[width=1\linewidth]{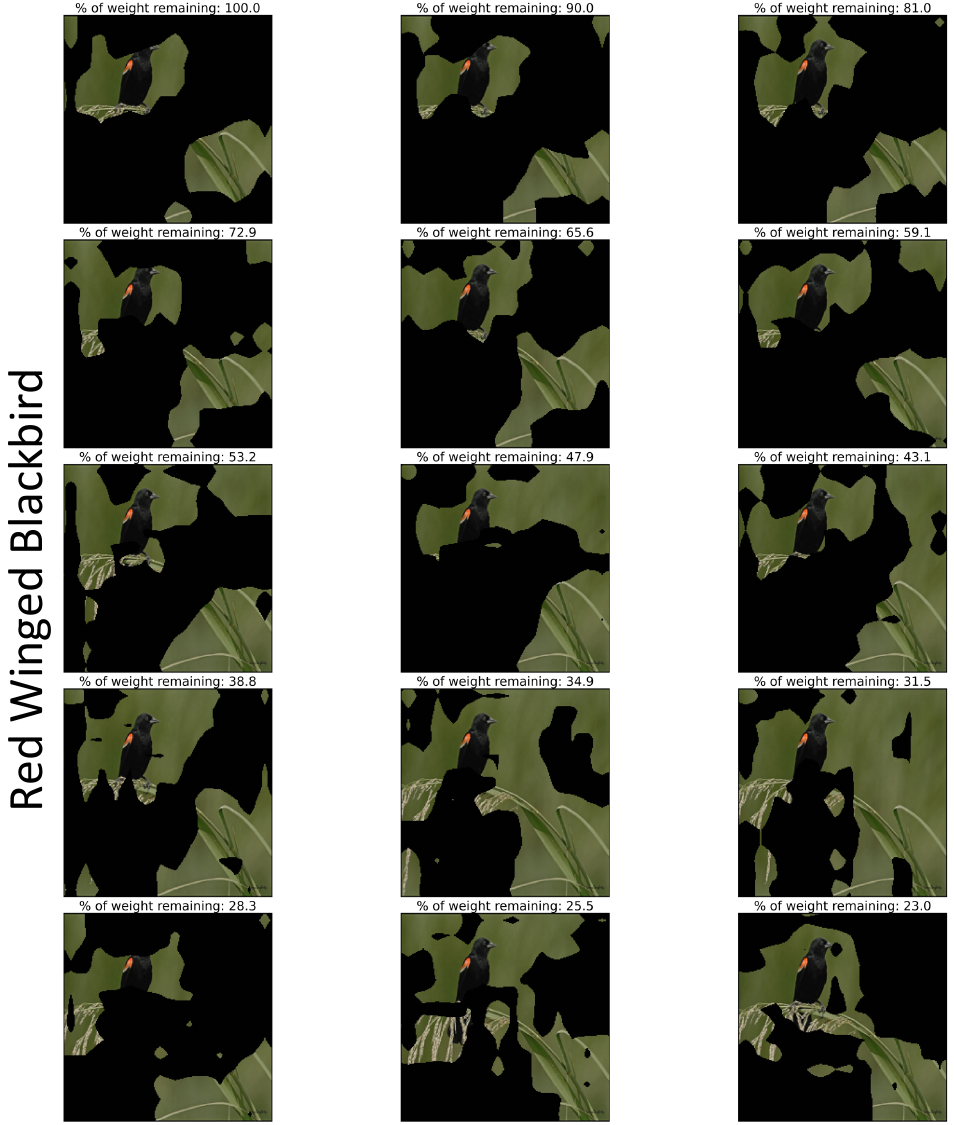}
\caption{Grad-CAM outputs of ``Red Winged Blackbird'' for various pruned models.}
\label{fig:grad_cam_cub}
\end{figure*}

\end{document}